\documentclass{article}

\usepackage{arxiv}

\usepackage[utf8]{inputenc} 
\usepackage[T1]{fontenc}    
\usepackage{hyperref}       
\usepackage{url}            
\usepackage{booktabs}       
\usepackage{amsfonts}       
\usepackage{nicefrac}       
\usepackage{microtype}      
\usepackage{graphicx}
\usepackage{natbib}
\usepackage{amsmath}
\usepackage{doi}

\usepackage[table]{xcolor}

\title{EMAGN: Efficient Multi-Attention Graph Network via Learned Clustering for Scalable Traffic Forecasting}

\author{%
  \textbf{Mingxing Xu}\textsuperscript{1*},\quad
  \textbf{Rakesh Chowdary Machineni}\textsuperscript{2*},\quad
  \textbf{Ke Liu}\textsuperscript{3*},\quad
  \textbf{Xi Cheng}\textsuperscript{4,\,\textdagger} \\[0.35em]
  \textbf{Chengqi Lu}\textsuperscript{5},\quad
  \textbf{Xin Hu}\textsuperscript{2},\quad
  \textbf{Lyuhao Chen}\textsuperscript{6},\quad
  \textbf{Xiangyu Li}\textsuperscript{7},\quad
  \textbf{Junwei You}\textsuperscript{8},\quad
  \textbf{Oliver Gao}\textsuperscript{4} \\[0.7em]
  \normalfont\small\itshape
  \textsuperscript{1}Shanghai Jiao Tong University \quad
  \textsuperscript{2}University of Michigan, Ann Arbor \quad
  \textsuperscript{3}University of California, Berkeley \\
  \normalfont\small\itshape
  \textsuperscript{4}Cornell University \quad
  \textsuperscript{5}Technische Universit\"at Dresden \\
  \normalfont\small\itshape
  \textsuperscript{6}Carnegie Mellon University \quad
  \textsuperscript{7}The University of Texas at Austin \quad
  \textsuperscript{8}University of Wisconsin--Madison
}

\date{}

\renewcommand{\shorttitle}{EMAGN: Linear-Complexity Attention for Traffic Forecasting}

\hypersetup{
  pdftitle={EMAGN: Efficient Multi-Attention Graph Network via Learned Clustering for Scalable Traffic Forecasting},
  pdfsubject={cs.LG, cs.AI},
  pdfauthor={Mingxing Xu, Rakesh Chowdary Machineni, Ke Liu, Xi Cheng, Chengqi Lu, Xin Hu, Lyuhao Chen, Xiangyu Li, Junwei You},
  pdfkeywords={Traffic forecasting, Self-attention, Graph neural networks, Linear complexity, Spatio-temporal modeling},
}

\newcommand\blfootnote[1]{%
  \begingroup
  \renewcommand\thefootnote{}%
  \footnotetext{#1}%
  \endgroup
}

\begin{document}
\maketitle

\blfootnote{\textsuperscript{*}Equal Contributions}
\blfootnote{\textsuperscript{\textdagger}Corresponding author:
  \texttt{xc557@cornell.edu}}

\begin{abstract}
Traffic forecasting is highly challenging due to complex and 
nonlinear spatial and temporal dependencies. Self-attention mechanisms have been widely adopted to model dynamic and long-range dependencies, 
achieving state-of-the-art performance, but suffer from limited 
scalability due to quadratic computational and memory complexity. 
To address this, we propose an Efficient Multi-Attention Graph Network 
(EMAGN) that linearises the spatial attention mechanism itself, inspired by the theory of fast high-dimensional Gaussian filtering. Two learned 
clustering matrices $C_k$ and $C_v$ adaptively group key and value 
vectors into $M$ super-clusters, reducing complexity from 
$\mathcal{O}(N^2d)$ to $\mathcal{O}(NMd)$ without sacrificing the 
flexibility of attention for dynamic dependency modelling.  Experimental results 
on PEMS-BAY and METR-LA show that EMAGN achieves accuracy within 
2.7--3.2\% MAE of full-attention GMAN while reducing training time by 
32\%, inference time by 38\%, and GPU memory by 58\%. Critically, at 
$K{=}16$ attention heads, full-attention GMAN runs out of memory on a 
standard 11\,GB GPU entirely while EMAGN continues to operate, 
demonstrating a categorical expansion of feasible model configurations. 
EMAGN also surpasses Linformer and Performer in both accuracy and 
efficiency within the same backbone, owing to its traffic-network-aware 
adaptive clustering.

\end{abstract}


\section{Introduction}
\label{intro}
\begin{figure}[t]
  \centering
  \includegraphics[width=0.96\textwidth]{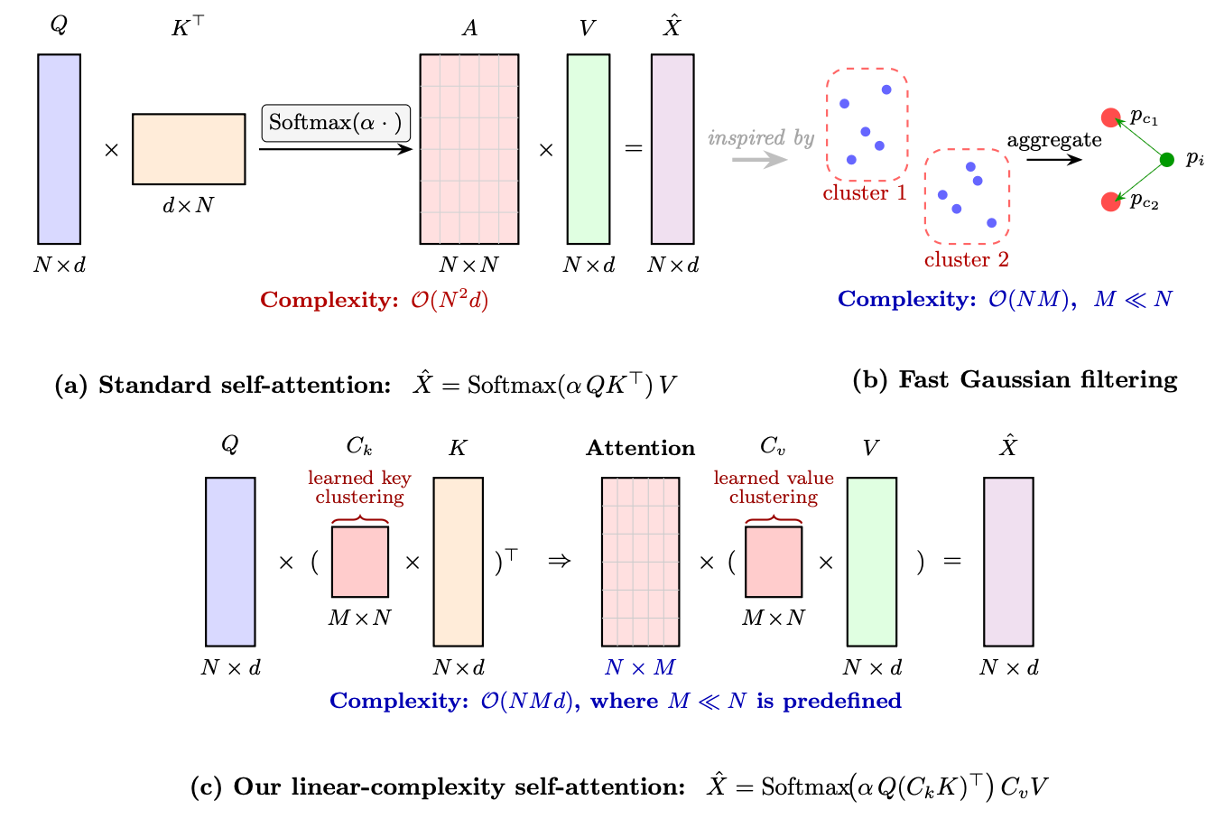}
  \caption{Illustration of the attention mechanisms. (a) Standard self-attention with $\mathcal{O}(N^2d)$ complexity. (b) Fast Gaussian filtering aggregates nearby positions into cluster centers. (c) Our linear-complexity self-attention uses learned clustering matrices $C_k$ and $C_v$ to compress keys and values, achieving $\mathcal{O}(NMd)$ complexity where $M \ll N$.}
\label{fig:attention_comparison}
\end{figure}

With the deployment of affordable traffic sensor technologies, traffic data are growing rapidly, bringing us to the era of transportation big data. Intelligent Transportation Systems (ITS)~\cite{mori2015review} leverage this data for efficient urban traffic control and planning. Traffic forecasting, namely predicting future traffic conditions (e.g., speeds, volumes, and density) from historical observations plays a critical role in ITS. Classical statistical 
methods~\cite{liu2011discovering,lippi2013short} rely on stationarity assumptions that are violated by the nonlinear, non-stationary nature of traffic flow.
Deep learning approaches, from CNNs~\cite{ma2017learning} and RNNs~\cite{wu2016short} to graph neural networks~\cite{yu2018spatio,li2018dcrnn_traffic,wu2019graph}, have progressively improved forecasting accuracy by incorporating spatial structure.
Self-attention mechanisms~\cite{vaswani2017attention} have been widely adopted to model dynamic and long-range spatial-temporal dependencies~\cite{GMAN-AAAI2020,park2020st}, achieving state-of-the-art performance. However, these self-attention-based models suffer from limited scalability due to the quadratic computational and memory complexity of the self-attention mechanism. As illustrated in Fig.~\ref{fig:attention_comparison}, pairwise interactions across all elements must be computed, resulting in $\mathcal{O}(N^2)$ complexity. The challenge grows with the addition of multi-head attention and deeper architectures, though both are essential for high performance on large-scale networks. While recent methods such as STAEformer~\cite{liu2023spatio} and 
MLCAFormer~\cite{he2025spatio} continue to advance accuracy through 
architectural redesign, adaptive embeddings, hierarchical attention, 
propagation-delay modelling, the memory bottleneck of quadratic spatial 
attention itself remains unaddressed. Despite this well-known limitation, existing traffic forecasting benchmarks 
do not report GPU memory consumption in their experimental evaluations, 
making it impossible to assess the practical deployability of these models 
on resource-constrained hardware.

To alleviate this bottleneck, we propose an Efficient Multi-Attention Graph Network (EMAGN), achieving comparable forecasting performance while being much more efficient and scalable. A linear-complexity self-attention mechanism is developed to model spatial dependencies across elements. Specifically, we adaptively group and aggregate $N$ key vectors into $M$ super-key vectors (super-value vectors are generated accordingly), and then dependencies between query and super-key vectors are calculated, reducing the computational complexity from $\mathcal{O}(N^2)$ to $\mathcal{O}(NM)$, where $M$ is a predefined hyperparameter. As a result, complexity scales linearly with the size of the traffic network. Our idea is derived from the fast high-dimensional Gaussian filtering technique in classical signal processing~\cite{adams2010fast}. The key insight is that value vectors associated with nearby positions can be combined into a single super-value vector, significantly reducing computational cost while achieving near-exact filtering performance.
EMAGN builds upon GMAN~\cite{GMAN-AAAI2020} as its base architecture, since GMAN explicitly computes full $N \times N$ spatial attention across all sensor pairs, making the quadratic bottleneck most pronounced and the efficiency gain most measurable. To address the scalability problem, GMAN proposed a group-spatial attention mechanism that randomly partitions vertices into groups and computes only intra-group and inter-group attentions. However, random grouping does not guarantee performance, and more groups are needed to achieve desirable accuracy. In contrast, we learn the grouping adaptively during training to exploit the inherent spatial clustering within traffic networks. Empirically, we find that a small number of learned groups suffices for competitive performance.
The main contributions of this paper are as follows:
\begin{itemize}
    \item We propose a linear-complexity self-attention mechanism derived from fast high-dimensional Gaussian filtering, reducing spatial attention complexity from $\mathcal{O}(N^2d)$ to $\mathcal{O}(NMd)$ through learned adaptive clustering matrices $C_k$ and $C_v$ that are trained end-to-end with the forecasting objective.
    \item We develop EMAGN, an efficient multi-attention graph network for traffic forecasting that achieves accuracy within 2.7--3.2\% MAE of full-attention GMAN while reducing training time by 32\%, inference time by 38\%, and GPU memory by 58\% on METR-LA.
    \item We show that EMAGN expands the feasible configuration space of attention-based traffic forecasting: at $K{=}16$ attention heads, full-attention GMAN exceeds the memory of a standard 11\,GB GPU, while EMAGN continues to operate at 8{,}568\,MB.
    \item We provide the first direct empirical comparison of general linear 
attention mechanisms, Linformer~\cite{wang2020linformer} and 
Performer~\cite{choromanski2022rethinkingattentionperformers}, against full attention on 
standard traffic benchmarks, demonstrating that EMAGN simultaneously 
achieves better accuracy and better efficiency than both alternatives.
\end{itemize}

\section{Related Work}
\label{sec:related}

\subsection{Traffic flow forecasting}

Early deep learning approaches to traffic forecasting employ CNNs to extract spatial features from grid-converted traffic networks~\cite{ma2017learning,zhang2019flow}, but this conversion discards the irregular topology inherent to road graphs.
Graph neural networks (GNNs) overcome this limitation by operating directly on the network structure~\cite{scarselli2008graph,kipf2017semi}. STGCN~\cite{yu2018spatio} applies spectral graph convolutions, while DCRNN and GraphWaveNet~\cite{li2018dcrnn_traffic,wu2019graph} adopt diffusion convolutions on directed graphs to capture directional flow.
These models, however, rely on fixed adjacency matrices and cannot adapt to time-varying spatial dependencies.
Adaptive-graph methods such as AGCRN~\cite{bai2020adaptive} and MTGNN~\cite{wu2020connecting} address this by learning the graph structure from data, yet they still depend on local message-passing and may struggle to capture long-range spatial interactions.

Self-attention mechanisms offer a natural solution to these problems with graph structure data, as they model pairwise dependencies across all nodes regardless of graph distance~\cite{pan2019urban,GMAN-AAAI2020,park2020st,wang2020traffic}.
GMAN~\cite{GMAN-AAAI2020} introduces a graph multi-attention mechanism with group-based spatial attention;
PDFormer~\cite{jiang2023pdformer} incorporates propagation delay-aware attention;
STAEformer~\cite{liu2023spatio} shows that a vanilla Transformer with well-designed adaptive embeddings can achieve strong results;
and MLCAFormer~\cite{he2025spatio} captures hierarchical spatio-temporal dependencies via multi-level causal attention.
Despite their effectiveness, all these models inherit the $\mathcal{O}(N^2)$ cost of standard self-attention, which limits their scalability to large networks.
Recent work has begun to tackle this bottleneck.
BigST~\cite{han2024bigst} achieves linear complexity through linearised graph convolutions, and STGformer~\cite{wang2024stgformer} reduces cost via a single-layer high-order attention design.
However, BigST sacrifices the flexibility of attention by relying on graph convolutions, while STGformer still depends on pre-computed graph structures.
In this paper, we take a different approach: we derive a linear-complexity self-attention mechanism from fast high-dimensional Gaussian filtering that preserves the full modelling flexibility of attention while scaling linearly with network size.

\subsection{Efficient attention mechanisms}
The quadratic complexity of standard self-attention~\cite{vaswani2017attention} has motivated extensive research on efficient alternatives, which can be broadly grouped into three families.
\textbf{Sparse attention} methods restrict the attention pattern to a subset of element pairs. Informer~\cite{zhou2021informer}, for example, selects dominant queries via a KL-divergence measure to achieve $\mathcal{O}(L\log L)$ complexity, but its sparse patterns are data-agnostic and require task-specific heuristics.
%
%
\textbf{Low-rank and kernel approximation} methods exploit the low-rank 
structure of the attention matrix. Linformer~\cite{wang2020linformer} projects 
keys and values into a lower-dimensional space, Performer~\cite{choromanski2022rethinkingattentionperformers} 
approximates the softmax kernel with positive orthogonal random features, and 
FMMformer~\cite{nguyen2021fmmformer} decomposes attention into near-field and 
far-field components inspired by the fast multipole method. While these methods 
achieve theoretical linear complexity, they apply fixed or random projection 
strategies that treat all sequence positions uniformly, without regard to the 
underlying data structure. In the traffic forecasting setting, where sensors 
exhibit strong geographic clustering and spatially structured dependencies, 
this position-agnostic treatment leads to suboptimal approximations of the 
full attention matrix.
\textbf{Clustering and inducing-point} methods are conceptually closest to our work. Set Transformer~\cite{lee2019set} introduces learnable inducing points to reduce pairwise computations, while Nystr\"omformer~\cite{xiong2021nystromformer} reconstructs the attention matrix from a subset of landmark points via the Nystr\"om approximation. However, these methods rely on generic clustering or random sampling strategies that are agnostic to the underlying data structure.

Despite the advances of efficient attention mechanisms, their direct application to spatio-temporal graph data for traffic forecasting remains limited. Traffic data also shows unique characteristics (e.g., irregular graph topology, jointly and dynamically evolving spatial and temporal dimensions) that general-purpose methods do not explicitly address. We bridge this gap by deriving a linear-complexity self-attention mechanism from the theory of fast high-dimensional Gaussian filtering, where the clustering matrices are learned end-to-end to adapt to the spatial structure of traffic networks.

\section{Proposed Model}
\label{sec:model}

We first derive a linear-complexity self-attention mechanism from the theory of fast high-dimensional Gaussian filtering, then develop the full EMAGN architecture for traffic forecasting.

\subsection{Linear-Complexity Self-Attention}
\label{sec:linear_attention}


Standard self-attention computes $\hat{X} = \text{Softmax}(\alpha QK^T)V$ 
with $\mathcal{O}(N^2d)$ complexity~\cite{vaswani2017attention}, where 
$Q, K, V$ are linearly projected from the input. This quadratic cost is 
structurally analogous to naive high-dimensional Gaussian filtering:
\begin{equation}\label{eq1}
\hat{v}_i = \sum_j e^{\frac{\|p_i - p_j\|^2}{2}} v_j,
\end{equation}
where $p_i$ and $v_i$ are high-dimensional positions and value vectors, 
which also requires all-pairs evaluation. Fast filtering algorithms~\cite{adams2010fast} reduce this to linear cost by grouping nearby positions into $M$ clusters and pre-aggregating their value contributions, as displayed in Fig.~\ref{fig:attention_comparison}. Specifically, given cluster centre $p_c$, the contribution of all elements in a cluster to a query $p_i$ is approximated via a truncated Taylor expansion as
\begin{equation}\label{eq3}
\textstyle\sum_je^{{\|p_i-p_j\|^2}/{2}}{v_j}\;\approx\; e^{{\|p_i-p_c\|^2}/{2}}\,P_r(\|p_i-p_c\|^2),
\end{equation}
where $P_r$ aggregates the $r$-th order Taylor terms and can be precomputed per cluster, so that each query attends only to $M \ll N$ cluster summaries rather than all $N$ elements.

Self-attention can be viewed as a generalisation of Gaussian filtering with directed dependencies and a scaled-product kernel rather than a Gaussian kernel. Following this analogy, we replace the hand-crafted clustering of classical filtering with two learnable clustering matrices $C_k, C_v \in \mathbb{R}^{M \times N}$ that are optimised end-to-end, allowing the grouping to adapt to the spatial structure of traffic networks. The resulting linear-complexity self-attention is
\begin{equation}
\hat{X} = \mathrm{Softmax}(\alpha Q (C_k K)^T) C_v V.
\label{eq:linear_attention}
\end{equation}
The attention matrix becomes $A = \alpha Q (C_k K)^T \in \mathbb{R}^{N \times M}$, reducing complexity from $\mathcal{O}(N^2 d)$ to $\mathcal{O}(NMd)$ (see Fig.~\ref{fig:attention_comparison}). Since $M$ is predefined and independent of $N$, computation scales linearly with network size. This mechanism can be replicated across multiple heads for richer representational capacity.

\subsection{EMAGN for Traffic Forecasting}
\label{sec:emagn}

\paragraph{Problem formulation.}
We denote $\mathcal{G}(\mathcal{V}, \mathcal{E}, W)$ as the input graph representing a traffic network, where $\mathcal{V}$ represents the nodes set with $N$ sensors, $\mathcal{E}$ is the set of edges reflecting physical connectivity between sensors and $W \in \mathbb{R}^{N \times N}$ is the adjacency matrix which is usually constructed with the Euclidean distances between sensors via Gaussian kernel. Formally, given $P$ historical traffic speeds 
$[v^{\tau-P+1}, \cdots, v^{\tau}] \in \mathbb{R}^{P \times N \times 1}$ 
observed by the $N$ sensors in the traffic network $\mathcal{G}$, a traffic 
forecasting model $\mathcal{F}$ is learned to predict the next $T$ time step 
traffic speeds $[\hat{v}^{\tau+1}, \cdots, \hat{v}^{\tau+T}]$ as
\begin{equation}
\hat{v}^{\tau+1}, \cdots, \hat{v}^{\tau+T} = \mathcal{F}(v^{\tau-P+1}, 
\cdots, v^{\tau}; \mathcal{G}).
\label{eq:problem}
\end{equation}


\begin{figure}[t]
\centering
\includegraphics[width=.75\textwidth]{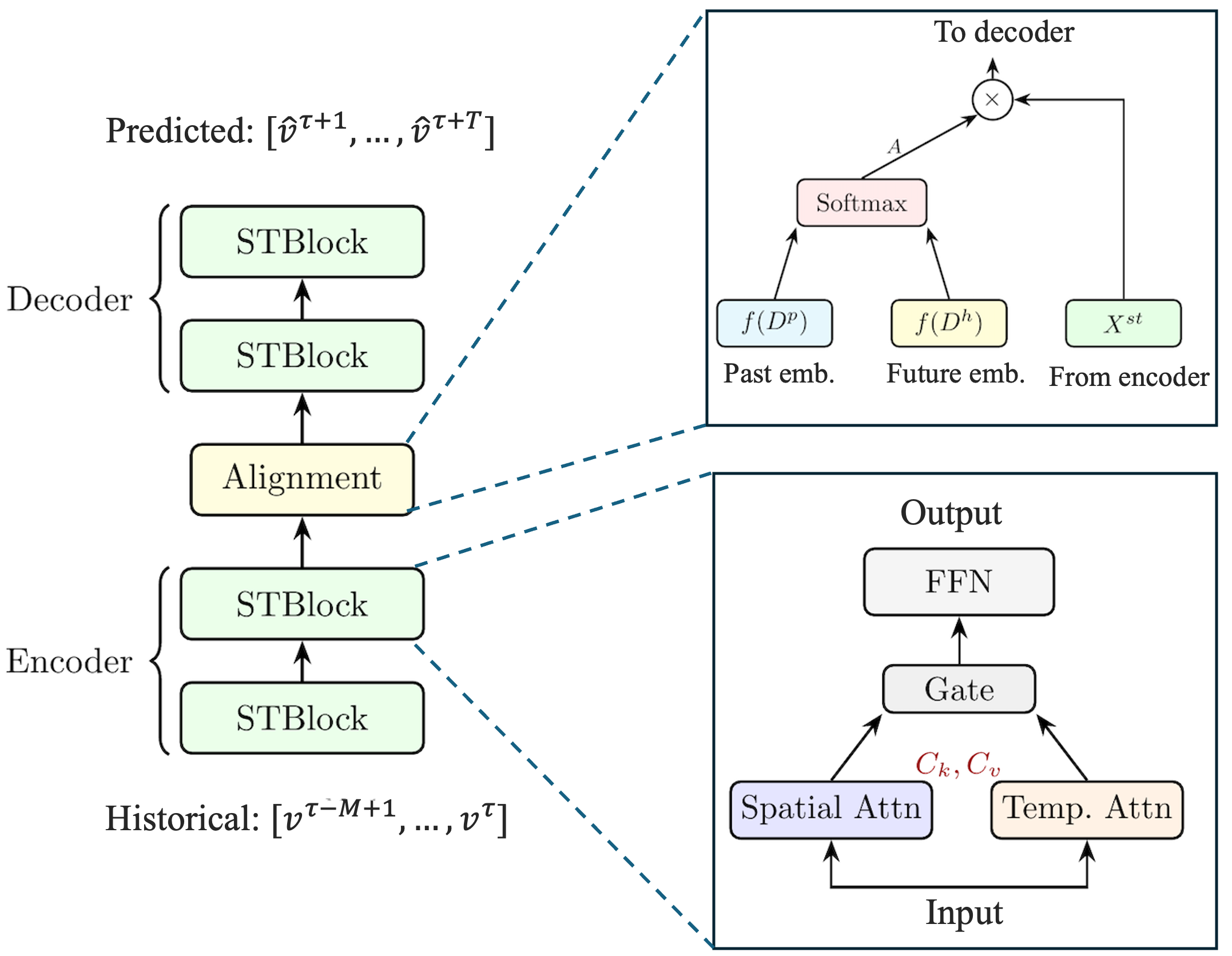}
\caption{The architecture of the proposed EMAGN. Left: overall encoder-decoder architecture with stacked STBlocks and an alignment module. Top right: the alignment module computes attention matrix $A$ from past and future embeddings, then transforms encoder output $X^{st}$ for the decoder. Bottom right: parallel fusion STBlock where spatial and temporal attention are computed independently and merged via a gated mechanism.}
\label{fig:architecture}
\end{figure}

\paragraph{Overall architecture.}
The architecture of the proposed model is illustrated in Fig.~\ref{fig:architecture}, consisting of an encoder and decoder. An alignment module that models the dynamical dependencies between historical and future time steps is also adopted. The STBlock consists of two modules: one for modeling spatial dependencies and another for modeling temporal dependencies. Spatial and temporal features are extracted independently within each STBlock and then fused using a gate. STBlocks are stacked to capture deep spatial and temporal features. To achieve spatial embedding, the node2vec algorithm~\cite{grover2016node2vec} is used to learn the embeddings $D_s$. For temporal embedding, we encode the day of the week and the time step in a day using one-hot encoding. These are then concatenated into a single vector, denoted as $D_t$. The spatial and temporal embeddings are both projected into a fixed-dimensional vector $D \in \mathbb{R}^{d_e}$ using two-layer fully connected networks. The learned spatial and temporal embeddings are concatenated with each input feature vector as positional embeddings. 

\paragraph{Spatial dependencies modeling.}
We propose adopting a linear-complexity self-attention mechanism to model spatial dependencies. For simplicity, we will use the example of time step $t_i$ for notation. Especially, for graph representations $X_{t_i} \in \mathbb{R}^{N \times (d_s + d_e)}$ that is either traffic speeds information or learned representation from the previous layer augmented with spatial-temporal embedding. Three linear projections $W_q^s \in \mathbb{R}^{(d_s+d_e) \times d_q}$, $W_k^s \in \mathbb{R}^{(d_s+d_e) \times d_k}$ and $W_v^s \in \mathbb{R}^{(d_s+d_e) \times d_v}$ and two clustering matrices $C_k \in \mathbb{R}^{M \times N}$ and $C_v \in \mathbb{R}^{M \times N}$ are learned. Formally, the updated graph features could be written as
\begin{equation}
\hat{X}_{t_i} = \mathrm{Softmax}\!\left(\frac{1}{\sqrt{d_k}} X_{t_i} W_q^s (C_k X_{t_i} W_k^s)^T\right) C_v X_{t_i} W_v^s.
\label{eq:spatial}
\end{equation}
In Eq.~(\ref{eq:spatial}), $Q = X_{t_i} W_q^s$, $K = C_k X_{t_i} W_k^s$ and $V = C_v X_{t_i} W_v^s$ are the learned query, super-key and super-value vectors, respectively. The key and value vectors are firstly adaptively grouped into $M$ clusters before modeling spatial dependencies. Compared to random grouping, we can achieve better performance with smaller groups. Multi-head attention can be further employed for various spatial patterns and the representations are further concatenated and fused with a two-layer fully connected network.

\paragraph{Temporal dependencies modeling.}
In this study, we also employ a self-attention mechanism for modeling dynamic and long-range temporal dependencies. Specifically, we consider time sequence features $H_{v_i} \in \mathbb{R}^{P \times (d_e + d_t)}$ of node $v_i$. The updated time features can be formally written as
\begin{equation}
\hat{H}_{v_i} = \mathrm{Softmax}\!\left(\frac{1}{\sqrt{d_k}} H_{v_i} W_q^t (H_{v_i} W_k^t)^T\right) H_{v_i} W_v^t
\label{eq:temporal}
\end{equation}
where $W_q^t \in \mathbb{R}^{(d_e+d_t) \times d_q}$, $W_k^t \in \mathbb{R}^{(d_e+d_t) \times d_k}$ and $W_v^t \in \mathbb{R}^{(d_e+d_t) \times d_v}$ are three linear projection matrices. As temporal sequences are fixed at $T = P =12$ in our setting, the quadratic 
cost of temporal attention is negligible---$\mathcal{O}(T^2 d)$ with 
$T \ll N$, and the dominant complexity of EMAGN remains 
$\mathcal{O}(NMd)$ in the spatial dimension. When the sequence grows much larger, linear complexity can be further considered. Similar to spatial dependencies modeling, multi-head attention can also be adopted and fused with fully connected networks.


\paragraph{Multi-step predictions.}
To model the dynamic temporal dependencies between past and future time steps, we employ an additional attention mechanism. By utilizing spatial and temporal embeddings for input and output time steps as features, we can effectively capture the dynamic temporal dependencies for different nodes. Formally, let $D_{v_i}^p \in \mathbb{R}^{T \times d_e}$ and $D_{v_i}^f \in \mathbb{R}^{T \times d_e}$ denote the past and future embeddings of node $v_i$, respectively, then the alignment matrix can be calculated with attention mechanism as
\begin{equation}
A = \mathrm{Softmax}\bigl(f(D_{v_i}^p)(f(D_{v_i}^f))^T\bigr),
\label{eq:alignment}
\end{equation}
where $A$ is the alignment matrix and $f$ are learnable linear projections. In our setting $P = T$, so both past and future embeddings share the same dimensionality. The output of encoder $X_{v_i}^{st}$ is then transformed with $A$ as $A X_{v_i}^{st}$ before fed into decoder.

\section{Experiments}
\label{sec:experiments}

We evaluate the proposed model on two real-world benchmark datasets: PEMS-BAY and METR-LA. As METR-LA is the more challenging benchmark, with higher speed variance (std 19.50 vs.\ 9.44 km/h) and stronger sensitivity to spatial structure, all ablation and scalability analyses are conducted on METR-LA unless otherwise stated.

\subsection{Datasets and Data Preprocessing}

PEMS-BAY consists of 6 months of speed data from 325 sensors in the Bay Area of California, starting from January 1st, 2017 through May 31st, 2017. METR-LA consists of traffic data from 207 sensor stations in the California state highway system during weekdays from May through June in 2012. The data statistics are presented in Table~\ref{tab:data}.
\begin{table}[t]
\caption{Dataset statistics. \# denotes the number of; std denotes the standard deviation of speeds.}\label{tab:data}
\centering
\begin{tabular}{lcccc}
\toprule
Dataset   & \#nodes & \#time steps & Mean speed & Std \\
\midrule
PEMS-BAY  & 325     & 52{,}116     & 62.74 km/h & 9.44  \\
METR-LA   & 207     & 34{,}272     & 54.40 km/h & 19.50 \\
\bottomrule
\end{tabular}
\end{table}
Following the common data processing pipeline, traffic speeds are aggregated every five minutes and normalised with Z-score. Data are split into 70\% training, 10\% validation, and 20\% testing. The road topology is encoded as a graph adjacency matrix using pairwise road distances via Gaussian kernel:
\begin{equation}
    W_{ij}= \begin{cases}
    \exp\!\left(\tfrac{-d_{ij}^2}{\sigma^2}\right), & \exp\!\left(\tfrac{-d_{ij}^2}{\sigma^2}\right)>\epsilon\\
    0, & \text{otherwise,}
    \end{cases}
\end{equation}
where $d_{ij}$ is the distance between sensors $v_i$ and $v_j$, $\sigma$ is the standard deviation and $\epsilon$ controls the sparsity of the adjacency matrix.

\subsection{Evaluation Metrics and Baselines}

We report mean absolute error (MAE), root mean squared error (RMSE), and mean absolute percentage error (MAPE), respectively, at 15, 30, and 60 minute prediction horizons. Baselines include graph-based deep models: STGCN~\cite{yu2018spatio}, DCRNN~\cite{li2018dcrnn_traffic}, GraphWaveNet~\cite{wu2019graph}, MTGNN~\cite{wu2020connecting}, and STID~\cite{shao2022spatial}. Among Transformer based methods ($\dagger$), we compare against PDFormer~\cite{jiang2023pdformer}, STAEformer~\cite{liu2023spatio}, MLCAFormer~\cite{he2025spatio}, and GMAN~\cite{GMAN-AAAI2020}, the full-attention model that EMAGN directly improves upon. For GMAN we adopt the published code with default settings; all remaining baseline results are reported from their original papers.

\subsection{Experimental Settings and Model Configurations}
All experiments are conducted on an NVIDIA RTX 2080Ti GPU. Models are trained using the Adam optimiser with an initial learning rate of $10^{-3}$, decaying by a factor of 0.7 every 5 epochs, and weight decay $10^{-3}$. Early stopping is applied after 10 consecutive epochs without validation improvement. We use 12 historical steps to predict the next 15-, 30-, and 60-minutes. Model depth is set proportional to dataset complexity, with PEMS-BAY using 
$L=1$ STBlock and METR-LA using $L=5$ STBlocks, both with $K=8$ attention heads and 8 channels per head. Unless stated otherwise, EMAGN uses $M=4$ clusters for METR-LA, determined by the ablation in Section~\ref{sec:ablation_clusters}.

\begin{table}[!t]
\caption{MAE, RMSE, and MAPE (\%) on PEMS-BAY and METR-LA. Top-2 results per column in \textbf{bold}. $\dagger$ denotes Transformer-based methods.}\label{tab:pems}
\centering
\setlength{\tabcolsep}{2.5pt}
\footnotesize
\begin{tabular}{l ccc ccc ccc}
\toprule
 & \multicolumn{3}{c}{15 min} & \multicolumn{3}{c}{30 min} & \multicolumn{3}{c}{60 min} \\
\cmidrule(lr){2-4}\cmidrule(lr){5-7}\cmidrule(lr){8-10}
Model & MAE & RMSE & MAPE & MAE & RMSE & MAPE & MAE & RMSE & MAPE \\
\midrule
\multicolumn{10}{l}{\textit{PEMS-BAY}} \\
\midrule
STGCN    & 1.36 & 2.96 & 2.90 & 1.81 & 4.27 & 4.17 & 2.49 & 5.69 & 5.79 \\
DCRNN    & 1.38 & 2.95 & 2.90 & 1.74 & 3.97 & 3.90 & 2.07 & 4.74 & 4.90 \\
GWNet    & \textbf{1.30} & \textbf{2.74} & \textbf{2.73} & 1.63 & 3.70 & 3.67 & 1.95 & 4.52 & \textbf{4.43} \\
MTGNN    & 1.33 & 2.80 & 2.81 & 1.66 & 3.77 & 3.75 & 1.95 & 4.50 & 4.62 \\
STID     & 1.31 & 2.79 & 3.78 & 1.64 & 3.73 & 3.73 & 1.91 & 4.42 & 4.55 \\
PDFormer$\dagger$    & 1.32 & 2.83 & 2.78 & 1.64 & 3.79 & 3.71 & 1.91 & 4.43 & 4.51 \\
STAEformer$\dagger$  & 1.31 & 2.78 & 2.76 & \textbf{1.62} & \textbf{3.68} & \textbf{3.62} & 1.88 & \textbf{4.34} & \textbf{4.41} \\
MLCAFormer$\dagger$  & \textbf{1.28} & \textbf{2.73} & \textbf{2.66} & \textbf{1.59} & \textbf{3.63} & \textbf{3.50} & \textbf{1.86} & \textbf{4.30} & \textbf{4.30} \\
GMAN$\dagger$        & 1.35 & 2.93 & 2.90 & 1.63 & 3.76 & 3.70 & \textbf{1.87} & 4.35 & 4.41 \\
EMAGN (ours) & 1.39 & 3.01 & 2.99 & 1.68 & 3.84 & 3.82 & 1.92 & 4.43 & 4.54 \\
\midrule
\rowcolor{gray!8} \multicolumn{10}{l}{\textit{METR-LA}} \\
\midrule
\rowcolor{gray!8} STGCN    & 2.88 & 5.74 & 7.62 & 3.47 & 7.24 & 9.57 & 4.59 & 9.40 & 12.70 \\
\rowcolor{gray!8} DCRNN    & 2.77 & 5.38 & 7.30 & 3.15 & 6.45 & 8.80 & 3.60 & 7.59 & 10.50 \\
\rowcolor{gray!8} GWNet    & 2.69 & 5.15 & 6.90 & 3.07 & 6.22 & 8.37 & 3.53 & 7.37 & 10.01 \\
\rowcolor{gray!8} MTGNN    & 2.69 & 5.16 & 6.89 & 3.05 & 6.13 & 8.16 & 3.47 & 7.21 & \textbf{9.70} \\
\rowcolor{gray!8} STID     & 2.82 & 5.53 & 7.75 & 3.19 & 6.57 & 9.39 & 3.55 & 7.55 & 10.95 \\
\rowcolor{gray!8} PDFormer$\dagger$    & 2.83 & 5.45 & 7.77 & 3.20 & 6.46 & 9.19 & 3.62 & 7.47 & 10.91 \\
\rowcolor{gray!8} STAEformer$\dagger$  & \textbf{2.65} & \textbf{5.11} & \textbf{6.85} & \textbf{2.97} & \textbf{6.00} & \textbf{8.13} & \textbf{3.34} & \textbf{7.02} & \textbf{9.70} \\
\rowcolor{gray!8} MLCAFormer$\dagger$  & \textbf{2.62} & \textbf{5.05} & \textbf{6.72} & \textbf{2.93} & \textbf{5.96} & \textbf{7.95} & \textbf{3.30} & \textbf{6.97} & \textbf{9.47} \\
\rowcolor{gray!8} GMAN$\dagger$        & 2.77 & 5.42 & 7.19 & 3.08 & 6.29 & 8.38 & 3.43 & 7.20 & 9.78 \\
\rowcolor{gray!8} EMAGN (ours) & 2.91 & 5.79 & 7.98 & 3.23 & 6.70 & 9.10 & 3.54 & 7.52 & 10.27 \\
\bottomrule
\end{tabular}
\end{table}


\begin{table}[t]
\caption{Comparison of linear attention mechanisms on METR-LA at the
60-minute prediction horizon ($L=5$, $K=8$, $M=4$).}\label{tab:linear_metr}
\centering
\begin{tabular}{l ccc ccc}
\toprule
 & \multicolumn{3}{c}{60 min} & \multicolumn{3}{c}{Efficiency} \\
\cmidrule(lr){2-4}\cmidrule(lr){5-7}
Model & MAE & RMSE & MAPE & Train (s) & Infer (s) & Mem (MB) \\
\midrule
GMAN      & 3.43 & 7.20 & 9.78 & 721.5 & 36.9 & 10666 \\
\textbf{EMAGN (ours)}     & \textbf{3.54} & \textbf{7.52} & \textbf{10.27} & \textbf{489.4} & \textbf{22.7} & \textbf{4472} \\
Linformer & 3.64 & 7.63 & 10.39 & 495.4 & 24.7 & 4562 \\
Performer & 3.62 & 7.62 & 10.78 & 496.6 & 24.7 & 4562 \\
\bottomrule
\end{tabular}
\end{table}
\subsection{Experimental Results}
Table~\ref{tab:pems} reports MAE, RMSE, and MAPE at the 15-, 30-, and 60-minute horizons on PEMS-BAY and METR-LA, respectively. We design the numerical experiments around three questions: (i)~how does EMAGN compare in absolute accuracy ? (ii)~what accuracy--efficiency trade-off does it achieve? and (iii)~does its learned-clustering mechanism outperform other linear attention designs?

\subsubsection*{Accuracy comparison.}
Quadratic-attention Transformers (i.e., MLCAFormer and STAEformer) achieve the highest overall accuracy on both benchmarks, benefiting from full $\mathcal{O}(N^2)$ pairwise interactions, followed by graph-based models (GraphWaveNet, MTGNN) and the full-attention GMAN.
EMAGN, operating at $\mathcal{O}(NM)$ complexity, incurs a moderate accuracy gap relative to these $\mathcal{O}(N^2)$ methods yet remains firmly within the range of established baselines.
On METR-LA at the 60-minute horizon, EMAGN achieves an MAE of 3.54\%, placing it within 3.2\% of its full-attention parent GMAN (3.43\%) and on par with GraphWaveNet (3.53\%) and DCRNN (3.60\%).
On PEMS-BAY at 60\,min, the gap narrows to 2.7\% MAE (1.92\% vs.\ GMAN's 1.87\%), while EMAGN substantially outperforms DCRNN (2.07\%) and STGCN (2.49\%).
A noteworthy trend emerges across prediction horizons: the accuracy gap between EMAGN and GMAN consistently shrinks as the forecasting horizon increases---from 5.1\% relative MAE at 15\,min to 3.2\% at 60\,min on METR-LA, and from 3.0\% to 2.7\% on PEMS-BAY.
This suggests that EMAGN's learned clustering captures the dominant spatial dependencies most critical for long-range prediction, where modelling global structure matters more than fine-grained pairwise interactions.

\subsubsection*{Accuracy--efficiency trade-off.}
Relative to GMAN, the full-attention architecture from which EMAGN is derived, our model reduces training time by 32\%
($721.5 \to 489.4$\,s/epoch), inference time by 38\%
($36.9 \to 22.7$\,s), and peak GPU memory by 58\%
($10{,}666 \to 4{,}472$\,MB) on METR-LA
(Table~\ref{tab:cluster_metr}, row $M{=}4$ vs.\ Full). As
Section~\ref{sec:scalability} demonstrates, these savings enable
model configurations, such as $K{=}16$ attention heads, that cause full-attention GMAN to exceed the memory of a standard 11\,GB GPU, whereas EMAGN continues to operate at 8{,}568\,MB.
Quadratic-attention baselines that achieve higher accuracy
(STAEformer, MLCAFormer) inherit the same $\mathcal{O}(N^2)$
bottleneck and therefore face identical scalability limits. EMAGN thus occupies a distinct position on the accuracy--efficiency Pareto front: it is the only method that simultaneously achieves accuracy within the range of established graph-based baselines and guarantees
linear-complexity attention.

\subsubsection*{Comparison with linear attention alternatives.}
The efficiency gains above could, in principle, be obtained by any linear attention mechanism. To isolate the contribution of EMAGN's learned clustering design, we replace its attention module with Linformer~\cite{wang2020linformer} and Performer~\cite{choromanski2022rethinkingattentionperformers}, two widely adopted
linear-attention methods, within the identical GMAN backbone. All three linear variants use the same network depth ($L{=}5$), head count ($K{=}8$), and projection dimension ($m{=}4$), ensuring a controlled comparison. Table~\ref{tab:linear_metr} reports accuracy at 60\,min and efficiency on METR-LA. EMAGN outperforms both alternatives across all reported metrics. In accuracy, it surpasses Linformer by 0.10 MAE (2.7\% relative) and Performer by 0.08 MAE (2.2\%); the gap is especially pronounced in MAPE, where EMAGN (10.27\%) improves over Performer (10.78\%) by 4.7\% relative. In efficiency, EMAGN consumes 2\% less memory (4{,}472 vs.\ 4{,}562\,MB) and runs 8\% faster at inference (22.7 vs.\ 24.7\,s). The advantage stems from the fact that Linformer and Performer apply fixed or random projections that treat all positions uniformly, whereas EMAGN learns separate clustering matrices $C_k$ and $C_v$ end-to-end that can adapt to the spatial structure of the traffic network.

\subsection{Ablation Study}
\label{sec:ablation_clusters}
\subsubsection*{Influence of the number of clusters.}
A key hyperparameter in EMAGN is the cluster count $M$, which controls the trade-off between approximation quality and computational efficiency. Table~\ref{tab:cluster_metr} lists accuracy and resource consumption as $M$ varies from 2 to 128, benchmarked against full $\mathcal{O}(N^2)$ attention.

\begin{table}[t]
\caption{Effect of cluster size $M$ on METR-LA. ``Full'' is standard quadratic
self-attention ($L{=}5$, $K{=}8$, batch 16). \textbf{Bold} is the default.}
\label{tab:cluster_metr}
\centering
\setlength{\tabcolsep}{4pt}
\small
\begin{tabular}{@{}l ccc ccc ccc ccc@{}}
\toprule
 & \multicolumn{3}{c}{15 min} & \multicolumn{3}{c}{30 min}
 & \multicolumn{3}{c}{60 min} & \multicolumn{3}{c}{Efficiency} \\
\cmidrule(lr){2-4}\cmidrule(lr){5-7}\cmidrule(lr){8-10}\cmidrule(lr){11-13}
$M$ & MAE & RMSE & MAPE & MAE & RMSE & MAPE & MAE & RMSE & MAPE
    & Trn (s) & Inf (s) & Mem (MB) \\
\midrule
Full & 2.77 & 5.42 & 7.19 & 3.08 & 6.29 & 8.38 & 3.43 & 7.20 & 9.78
     & 721 & 36.9 & 10{,}666 \\
128  & 2.96 & 5.77 & 8.23 & 3.29 & 6.69 & 9.54 & 3.61 & 7.46 & 10.80
     & 619 & 30.2 & 8{,}543 \\
64   & 2.93 & 5.83 & 8.14 & 3.26 & 6.76 & 9.47 & 3.60 & 7.60 & 10.81
     & 593 & 28.7 & 7{,}728 \\
32   & 2.95 & 5.88 & 8.35 & 3.29 & 6.83 & 9.68 & 3.61 & 7.64 & 10.89
     & 567 & 27.2 & 7{,}728 \\
16   & 2.91 & 5.80 & 8.08 & 3.24 & 6.75 & 9.47 & 3.58 & 7.61 & 10.85
     & 541 & 25.7 & 6{,}100 \\
8    & 2.90 & 5.76 & 7.91 & 3.22 & 6.65 & 9.15 & 3.58 & 7.52 & 10.52
     & 515 & 24.2 & 5{,}286 \\
\textbf{4} & \textbf{2.91} & \textbf{5.79} & \textbf{7.98}
           & \textbf{3.23} & \textbf{6.70} & \textbf{9.10}
           & \textbf{3.54} & \textbf{7.52} & \textbf{10.27}
           & \textbf{489} & \textbf{22.7} & \textbf{4{,}472} \\
2    & 2.92 & 5.75 & 8.03 & 3.22 & 6.56 & 9.21 & 3.59 & 7.44 & 10.72
     & 478 & 21.9 & 4{,}250 \\
\bottomrule
\end{tabular}
\end{table}
Three findings are noteworthy. First, $M=4$ achieves accuracy essentially on par with $M=8$ (MAE 2.91 vs.\ 2.90 at 15 min; 3.54 vs.\ 3.58 at 60 min) while using 4,472\,MB versus 5,286\,MB, and both are far below full attention's 10,666\,MB. Second, larger cluster sizes ($M=64$, $M=128$) do not consistently improve over smaller ones and in several cases perform worse. This behaviour is consistent with the Gaussian filtering analogy: once the learned clusters capture the dominant spatial groupings in the traffic network, additional clusters provide diminishing approximation benefit while increasing compute. Third, and most significantly, the fact that $M=4$ matches or exceeds $M=128$ on accuracy confirms that traffic sensor graphs have inherent low-rank spatial structure. A small number of learned clusters suffices because sensors naturally group into coherent geographic corridors, a structure that EMAGN discovers end-to-end, and that generic low-rank projections (Linformer, Performer) cannot exploit. We set $M=4$ as the default for METR-LA.

\subsection{Scalability Analysis}
\label{sec:scalability}
Beyond the efficiency gains at existing configurations, EMAGN's linear complexity enables model configurations that are infeasible under full attention. We demonstrate this through three controlled scaling experiments on METR-LA.

\subsubsection*{Scalability with the Number of STBlocks $L$}

Table~\ref{tab:scale_L} shows resource consumption as more STBlocks are stacked. Both models scale roughly linearly in $L$, as expected. However, EMAGN's costs are consistently and substantially lower at every depth. Notably, even at $L=1$, EMAGN already consumes 58\% less memory (895\,MB vs.\ 2,133\,MB) and trains 23\% faster than GMAN, confirming the efficiency advantage is not a deep-model artefact but is present at every configuration. At $L=5$, EMAGN requires 489.4s training time and 4,472\,MB memory versus GMAN's 721.5s and 10,666\,MB, reductions of 32\% and 58\% respectively. Within a fixed memory budget, EMAGN can therefore deploy significantly deeper models than GMAN, enabling richer spatial-temporal feature extraction that is entirely inaccessible under full attention.

\begin{table}[t]
\caption{Scalability with STBlocks $L$ on METR-LA ($K=8$, $M=4$ for EMAGN).}\label{tab:scale_L}
\centering
\begin{tabular}{l ccc ccc}
\toprule
 & \multicolumn{3}{c}{GMAN (Full Attention)} & \multicolumn{3}{c}{EMAGN (Linear Attention)} \\
\cmidrule(lr){2-4}\cmidrule(lr){5-7}
$L$ & Train (s) & Infer (s) & Mem (MB) & Train (s) & Infer (s) & Mem (MB) \\
\midrule
1 & 172.4 & 8.7  & 2{,}133  & 133.0 & 6.2  & 895  \\
2 & 311.1 & 15.8 & 4{,}447  & 224.0 & 11.1 & 1{,}850 \\
3 & 446.8 & 22.8 & 6{,}495  & 315.4 & 16.0 & 2{,}760 \\
4 & 579.4 & 29.7 & 8{,}543  & 406.8 & 20.9 & 3{,}680 \\
5 & 721.5 & 36.9 & 10{,}666 & 489.4 & 22.7 & 4{,}472 \\
\bottomrule
\end{tabular}
\end{table}

\subsubsection*{Scalability with the Number of Attention Heads $K$}

\begin{figure*}[t]
    \centering
    \includegraphics[width=\textwidth]{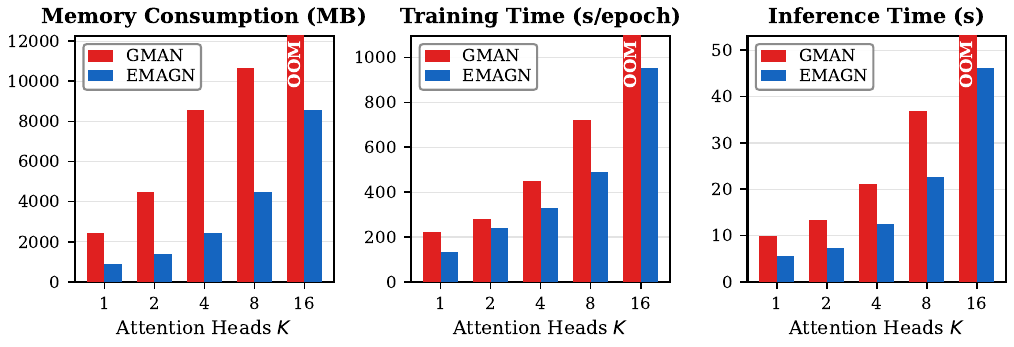}
    \caption{Scalability with number of attention heads $K$ on METR-LA ($L=5$, $M=4$ for EMAGN). \textbf{OOM}: out of memory on a single 11\,GB GPU.}
    \label{fig:scale_h}
\end{figure*}

Figure~\ref{fig:scale_h} reports the effect of varying the number of attention heads $K$. At $K{=}16$, GMAN exceeds the 11\,GB GPU memory (OOM), whereas EMAGN continues to operate at 954\,s training time and 8{,}568\,MB. At the shared configuration $K{=}8$, EMAGN consumes 58\% less memory and runs $1.6{\times}$ faster at inference, confirming that the efficiency advantage grows with the number of attention heads.

\subsubsection*{Scalability with Traffic Network Size $N$}

\begin{figure*}[!t]
  \centering
  \includegraphics[width=1.0\linewidth]{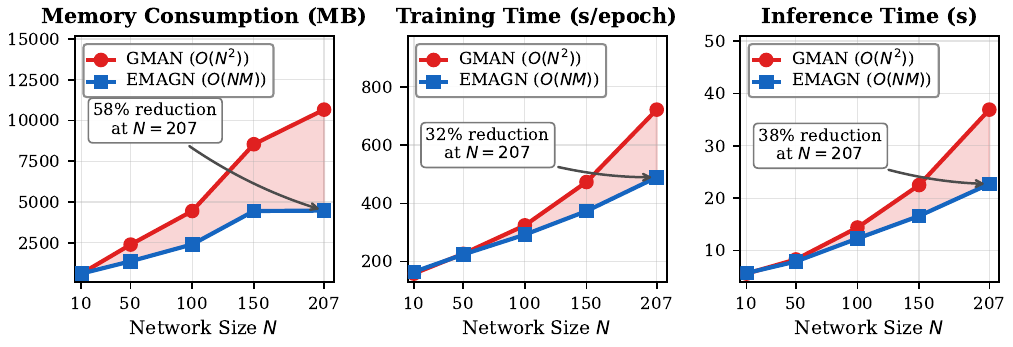}
  \caption{Scalability with traffic network size $N$ on METR-LA ($L=5$, $K=8$, $M=4$ for EMAGN).}
  \label{fig:figscalability}
\end{figure*}

Figure~\ref{fig:figscalability} evaluates how costs scale with network size $N$ by sub-sampling the METR-LA graph. At small $N$ ($N=10$), both models exhibit similar resource requirements since the quadratic term $N^2$ is negligible. As $N$ grows, the gap widens significantly and follows the predicted pattern: GMAN's costs grow steeply while EMAGN's costs follow a near-linear trajectory, directly validating the theoretical $\mathcal{O}(NM)$ complexity. At $N=207$, EMAGN requires 32\% less training time and 58\% less memory. For deployment on larger real-world networks, city-scale systems with thousands of sensors, this gap would grow substantially further, making full-attention models practically infeasible while EMAGN remains deployable.

Together, the three scalability analyses consistently confirm that EMAGN achieves its theoretical $\mathcal{O}(NM)$ advantage in practice across all three axes of scale. The efficiency gains are not merely cost reductions at existing configurations, they translate directly into new capabilities: deeper networks, richer multi-head configurations, and larger-scale deployments that full-attention models cannot support within standard hardware budgets.

\section{Conclusion}
\label{sec:conclusion}
In this paper, we proposed EMAGN, an efficient multi-attention graph network for traffic speed forecasting. Inspired by the high-dimensional Gaussian filtering, we developed a linear-complexity attention mechanism that replaces the $\mathcal{O}(N^2)$ pairwise interactions of standard self-attention with $\mathcal{O}(NM)$ operations through learned adaptive clustering. The EMAGN architecture stacks spatial-temporal blocks with parallel fusion of spatial and temporal attention, combined with a transform-attention alignment module for multi-step prediction.

Extensive experiments on PEMS-BAY and METR-LA demonstrate that EMAGN achieves accuracy closely competitive with the state-of-the-art full-attention model GMAN, with only a 2.7--3.2\% MAE gap at the 60-minute horizon, while reducing training time by 32\%, inference time by 38\%, and memory consumption by 58\%. Scalability experiments further show that EMAGN supports configurations (e.g., 16 attention heads) that exceed the memory capacity of full-attention models. Ablation studies confirm that as few as 4--8 learned clusters suffice to approximate full pairwise attention, validating the design principle derived from fast Gaussian filtering. A controlled comparison with Linformer and Performer further shows that learned clustering achieves better accuracy and efficiency than generic linear attention alternatives within the same backbone.

Future work includes extending the linear attention mechanism to temporal dependencies when dealing with longer historical sequences, exploring adaptive selection of cluster size $M$ during inference, and evaluating EMAGN on larger city-scale traffic networks.



\bibliographystyle{unsrtnat}
\bibliography{references}  






\end{document}